\documentclass{article}

     \PassOptionsToPackage{numbers, compress}{natbib}

     \usepackage[preprint]{neurips_2024}

\usepackage[utf8]{inputenc} 
\usepackage[T1]{fontenc}    
\usepackage{hyperref}       
\usepackage{url}            
\usepackage{booktabs}       
\usepackage{amsfonts}       
\usepackage{nicefrac}       
\usepackage{microtype}      
\usepackage{xcolor}         
\usepackage{bbm}
\usepackage{amsfonts}
\usepackage{siunitx}

\usepackage{subfigure}

\usepackage{amsmath}
\usepackage{amssymb}
\usepackage{mathtools}
\usepackage{amsthm}
\usepackage{dsfont}
\usepackage{soul}
\usepackage{enumitem}
\usepackage{arydshln}

\theoremstyle{plain}

\theoremstyle{definition}

\theoremstyle{remark}

\usepackage{etoc}
\etocdepthtag.toc{mtchapter}
\etocsettagdepth{mtchapter}{subsection}
\etocsettagdepth{mtappendix}{none}

\allowdisplaybreaks

\title{LAMP: Learnable Meta-Path Guided Adversarial Contrastive Learning for Heterogeneous Graphs}

\author{%
  Siqing Li \\
  University of New South Wales \\
  \texttt{siqing.li@unsw.edu.au} \\
  \And
  Jin-Duk Park \\
  Yonsei University \\
  \texttt{jindeok6@yonsei.ac.kr} \\
  \And
  Wei Huang\thanks{Corresponding author.} \\
  RIKEN AIP \\
  \texttt{wei.huang.vr@riken.jp} \\
  \And
  Xin Cao\footnotemark[1] \\
  University of New South Wales \\
  \texttt{xin.cao@unsw.edu.au} \\
  \And
  Won-Yong Shin \\
  Yonsei University \\
  \texttt{wy.shin@yonsei.ac.kr} \\
  \And
  Zhiqiang Xu \\
  Mohamed bin Zayed University of Artificial Intelligence \\
  \texttt{zhiqiangxu2001@gmail.com} \\
}

\begin{document}

\maketitle

\begin{abstract}
  Heterogeneous graph neural networks (HGNNs) have significantly propelled the information retrieval (IR) field. Still, the effectiveness of HGNNs heavily relies on high-quality labels, which are often expensive to acquire. This challenge has shifted attention towards Heterogeneous Graph Contrastive Learning (HGCL), which usually requires pre-defined meta-paths.
However, our findings reveal that meta-path combinations significantly affect performance in unsupervised settings, an aspect often overlooked in current literature. 
Existing HGCL methods have considerable variability in outcomes across different meta-path combinations, thereby challenging the optimization process to achieve consistent and high performance. In response, we introduce \textsf{LAMP} (\underline{\textbf{L}}earn\underline{\textbf{A}}ble \underline{\textbf{M}}eta-\underline{\textbf{P}}ath), a novel adversarial contrastive learning approach that integrates various meta-path sub-graphs into a unified and stable structure, leveraging the overlap among these sub-graphs. To address the denseness of this integrated sub-graph, we propose an adversarial training strategy for edge pruning, maintaining sparsity to enhance model performance and robustness. \textsf{LAMP} aims to maximize the difference between meta-path and network schema views for guiding contrastive learning to capture the most meaningful information. Our extensive experimental study conducted on four diverse datasets from the Heterogeneous Graph Benchmark (HGB) demonstrates that \textsf{LAMP} significantly outperforms existing state-of-the-art unsupervised models in terms of accuracy and robustness.
\end{abstract}

\section{Introduction}

Heterogeneous graphs characterized by diverse node and edge types are ubiquitous across various domains including social, academic, and user interaction networks. 
The use of heterogeneous graph neural networks (HGNNs) has surged in IR applications, ranging from search engines~\cite{DBLP:conf/mm/ChenWC00P22,guan2022personalized,yang2020biomedical} to recommendation systems~\cite{cai2023user,pang2022heterogeneous,sun2020multi,DBLP:conf/sigir/JiangLCW23,DBLP:conf/sigir/LiHDZZHM23} and question answering systems~\cite{feng2022multi,gao2022heteroqa,DBLP:conf/sigir/ChristmannRW23}.

HGNNs fall into two categories: Meta-path based models~\cite{HAN,HetGNN,MAGNN,SimpleHGNN}, converting HINs into homogeneous sub-graphs via predefined meta-paths, and Meta-path free models~\cite{GTN,RSHN,HGT,HetSANN,HGB,yu2022multiplex}, facilitating distinct information propagation along varied relations. These models have shown promising results but often require extensive labeling, posing challenges for large-scale IR tasks. Consequently, there has been a shift towards self-supervised learning (SSL) approaches, particularly in Heterogeneous Graph Contrastive Learning (HGCL)~\cite{DMGI,hdmi,HeCo,XGOAL,chen2023heterogeneous,zheng2022contrastive,zhu2022structure}.

In HGCL, a widely adopted approach involves the generation of multiple graph views via diverse data augmentation techniques, subsequently refining node representations through contrastive learning. Two principal categories of HGCL augmentations emerge: (1) the meta-path view~\cite{DMGI,XGOAL,hdmi}, which converts heterogeneous graphs into homogeneous sub-graphs according to selected meta-paths, and (2) the network schema view~\cite{HeCo,park2022crossview}, wherein the target nodes aggregate information from one-hop neighbors of varying node types. Distinctively, the network schema view imparts a localized perspective, while the meta-path view delivers a more expansive, higher-order perspective, connecting target nodes through meta-path instances that span multiple hops. However, recent studies have revealed that manually crafted augmentations, including the prevalent meta-path view, often fall short of achieving optimal results~\cite{zhu2021empirical,zhu2021deep,hussein2018meta}. This demonstrates a significant reliance on the specific combination of meta-paths chosen, which in turn, greatly affects the overall model performance.

\begin{figure}[tbp] 
\centering 
\includegraphics[width=0.60\linewidth]{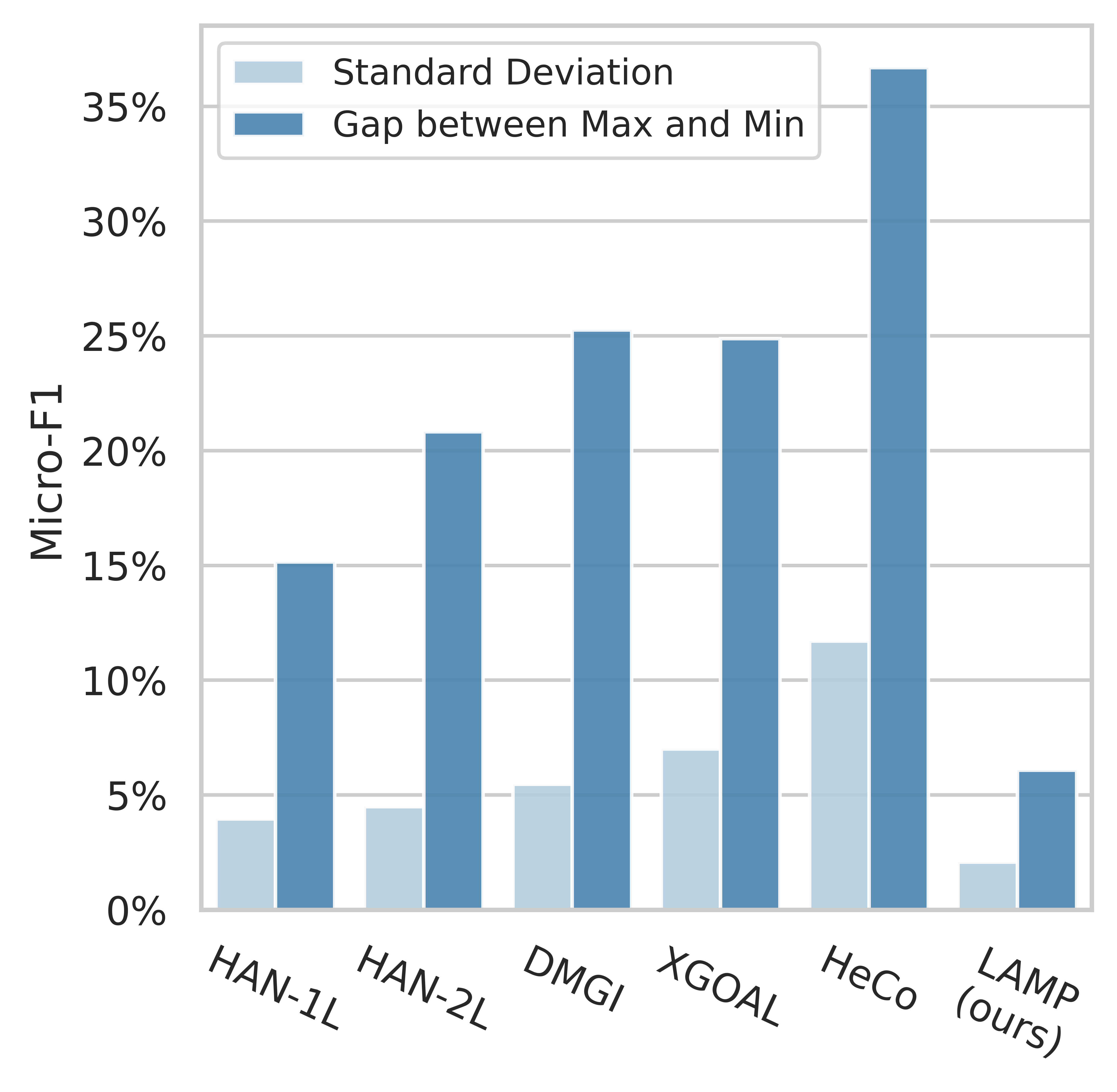} 
\caption{\small{Comparing performance variability in node classification and illustrating standard deviation and min-max gaps across HGNN models (supervised HAN with 1/2 layers, unsupervised XGOAL, HeCo, DMGI, and our \textsf{LAMP}) using varied meta-path combinations.}}
\label{fig:metapath_std} 
\end{figure}

In this study, we explore the relationship between the meta-path set selection and HGNN model performance on node classification, detailed in Section \ref{section:Empirical Observations}. Our findings illustrated in Figure~\ref{fig:metapath_std} reveal that the set of meta-paths selected crucially affect model performance, with all models showing at least a 5\% deviation across different combinations, especially pronounced in SSL models. Clearly, the identification of the optimal meta-path combination is crucial, yet presents considerable challenges due to:

\smallskip\noindent\textbf{(1) No Universal Meta-Path Combination}: Our research indicates the absence of a universally optimal meta-path combination among models, with effectiveness varying significantly (see Figure~\ref{fig:metapath_comb_ranking}). The optimal set for supervised models often underperforms in unsupervised scenarios, highlighting SSL's inherent complexity.

\noindent\textbf{(2) No use in Adding More Meta-paths}: Surprisingly, adding more meta-paths doesn't consistently lead to better performance. Although effective in supervised learning contexts as evidenced by SOTA methods~\cite{SimpleHGNN,10.1145/3539618.3591765}, this approach does not translate as effectively into SSL scenarios. Consequently, a straightforward greedy search for the optimal meta-path combination is inadequate in the SSL landscape.

\noindent\textbf{(3) No Downstream Task Labels in SSL}: SSL methods face a unique challenge in that they cannot employ downstream tasks to determine the most effective meta-path combinations, as these tasks are not applicable in unsupervised contexts. 

Addressing the issue in an unsupervised framework, our solution is to increase the robustness of HGCL models against diverse meta-path combinations. The existing models lack robustness primarily because each meta-path is treated as an independent channel, making changes in these channels potentially harmful to model stability. To address the overlooked issue of meta-path sensitivity, we present \textsf{LAMP} — a \underline{\textbf{L}}earn\underline{\textbf{A}}ble \underline{\textbf{M}}eta-\underline{\textbf{P}}ath guided adversarial contrastive learning model which aims at creating a stable meta-path view. It reduces dependency on specific meta-path combinations and achieves consistent performance, also simplifying the integration of a wide range of meta-paths. Furthermore, we enhance \textsf{LAMP} with adversarial training, a technique known to improve contrastive learning performance in homogeneous graphs.

\textsf{LAMP} proposes a new perspective in meta-path view construction by merging different meta-path sub-graphs into a unified structure. This results in a singular sub-graph that integrates nodes and edges from various meta-path sub-graphs. In this integrated sub-graph, each edge carries a one-hot-like encoding based on its meta-path instance, maintaining the semantic integrity of the original sub-graphs. This unified form ensures stability across various meta-path combinations, utilizing the overlaps between them. For instance, combining sub-graphs from {PAP,PSP,PAPAP} (refer to Figure~\ref{fig:HG} (b)) into one integrated sub-graph (Figure~\ref{fig:HG} (c)) retains the topological structure when modifying the combination, such as removing a meta-path, but with different edge encoding. This stability stems from the shared edges commonly found in heterogeneous graphs, as detailed in Section~\ref{section:Empirical Observations}, thereby significantly reducing variability between combinations and enhancing the model's robustness.

Nevertheless, As the number of meta-paths in the integrated sub-graph increases, so does its density, which may hinder performance since Graph Contrastive Learning (GCL) generally performs better with sparser structures\cite{zhu2021empirical}. In extreme cases, the integrated sub-graph might become too dense, resembling a complete graph, and lead to high computational cost. To address this, we apply an adversarial training method named \textsf{LMA} (Learnable Meta-path Guided Augmentation). Initially, \textsf{LMA} simplifies the graph by randomly removing edges. It then employs a learned edge-pruning approach, guided by node features and semantic information, to optimally refine the graph's structure. This process enhances both the model's efficiency and its robustness. The edge encoding is combined with a learnable weight vector to represent the importance of different meta-paths. \textsf{LMA}'s goal is to create a significant distinction between the network schema and meta-path views, allowing the HGCL framework to effectively extract the most meaningful knowledge. This approach follows the adversarial training model common in graph contrastive learning. Our comprehensive experiments on four HGB~\cite{HGB} real-world datasets demonstrate that \textsf{LAMP} not only outperforms current SOTA baselines but also greatly improves robustness.

\section{Preliminary}

\begin{figure}[tbp] 
\centering 
\includegraphics[width=0.8\linewidth]{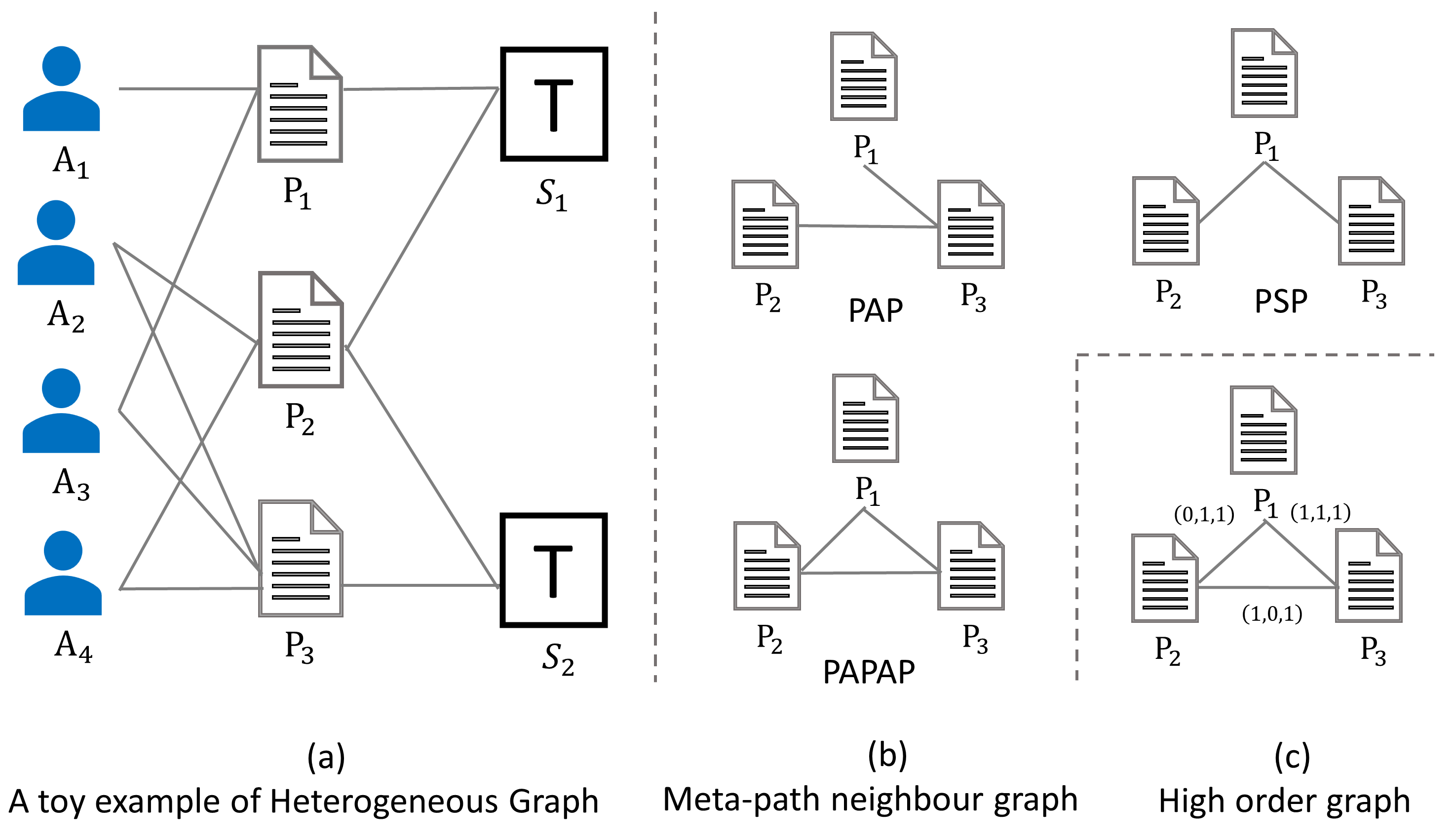} 
\caption{\small{A simplistic toy example derived from the ACM dataset: (a) Illustrates a Heterogeneous Graph. (b) Demonstrates three distinct meta-path sub-graphs associated with their respective meta-paths: PAP, PSP, and PAPAP. (c) Displays an integrated meta-path sub-graph that aggregates all the meta-path sub-graphs; its edge type embedding indicates which meta-paths are involved in each edge.
}}
\label{fig:HG} 
\end{figure}

\textbf{\emph{Definition 1. Heterogeneous Information Network (HIN).}} 
A HIN is a network \( \mathcal{G} = (\mathcal{V}, \mathcal{E}, \mathcal{A}, \mathcal{R}, \mathcal{\theta}, \mathcal{\phi}) \), where \( \mathcal{V} \) and \( \mathcal{E} \) represent the sets of nodes and edges, respectively. The network is associated with a node type mapping function \( \mathcal{V} \rightarrow \mathcal{A} \) and an edge type mapping function \( \mathcal{\phi}: \mathcal{E} \rightarrow \mathcal{R} \). Here, \( \mathcal{A} \) and \( \mathcal{R} \) represent the sets of object and link types, respectively, with the constraint \( |\mathcal{A}| + |\mathcal{R}| > 2 \).\\
\textbf{\emph{Definition 2. Meta-path.}} 
A meta-path \( \mathcal{P} \) is a structural pattern connecting different node types, represented as 
\[ A_1 \xrightarrow{R_1} A_2 \xrightarrow{R_2} A_3 \cdots \xrightarrow{R_l} A_{l+1} \] 
(abbreviated as \( A_1 A_2 \dots A_{l+1} \)), which describes a composite relation \( R = R_1 \circ R_2 \circ \dots \circ R_l \) between node types \( A_1 \) and \( A_{l+1} \), where \( \circ \) represents the composition operator on relations. Paths in \( \mathcal{G} \) that follow the pattern of \( \mathcal{P} \) \ are termed as meta-path instances.\\
\textbf{\emph{Definition 3. Meta-path Sub-Graph.}} 
Given a meta-path \( \mathcal{P} \), the nodes in \( \mathcal{G} \) can be re-connected to form a meta-path sub-graph \( \mathcal{G}_\mathcal{P} \). An edge \( e \rightarrow v \) exists in \( \mathcal{G}_\mathcal{P} \) if and only if there's at least one path (a meta-path instance) between \( u \) and \( v \) following the meta-path \( \mathcal{P} \) in the original graph \( \mathcal{G} \). For instance, Figure~\ref{fig:HG} (b) illustrates three meta-path sub-graphs derived from the HIN in Figure~\ref{fig:HG} (a). PAP indicates two papers authored by the same individual, while PSP signifies two papers related to the same subject. As meta-paths combine multiple relations, meta-path sub-graphs encapsulate high-order structures.

\section{Empirical Observations}
\label{section:Empirical Observations}

\begin{figure}[tbp] 
\centering 
\includegraphics[width=0.8\linewidth]{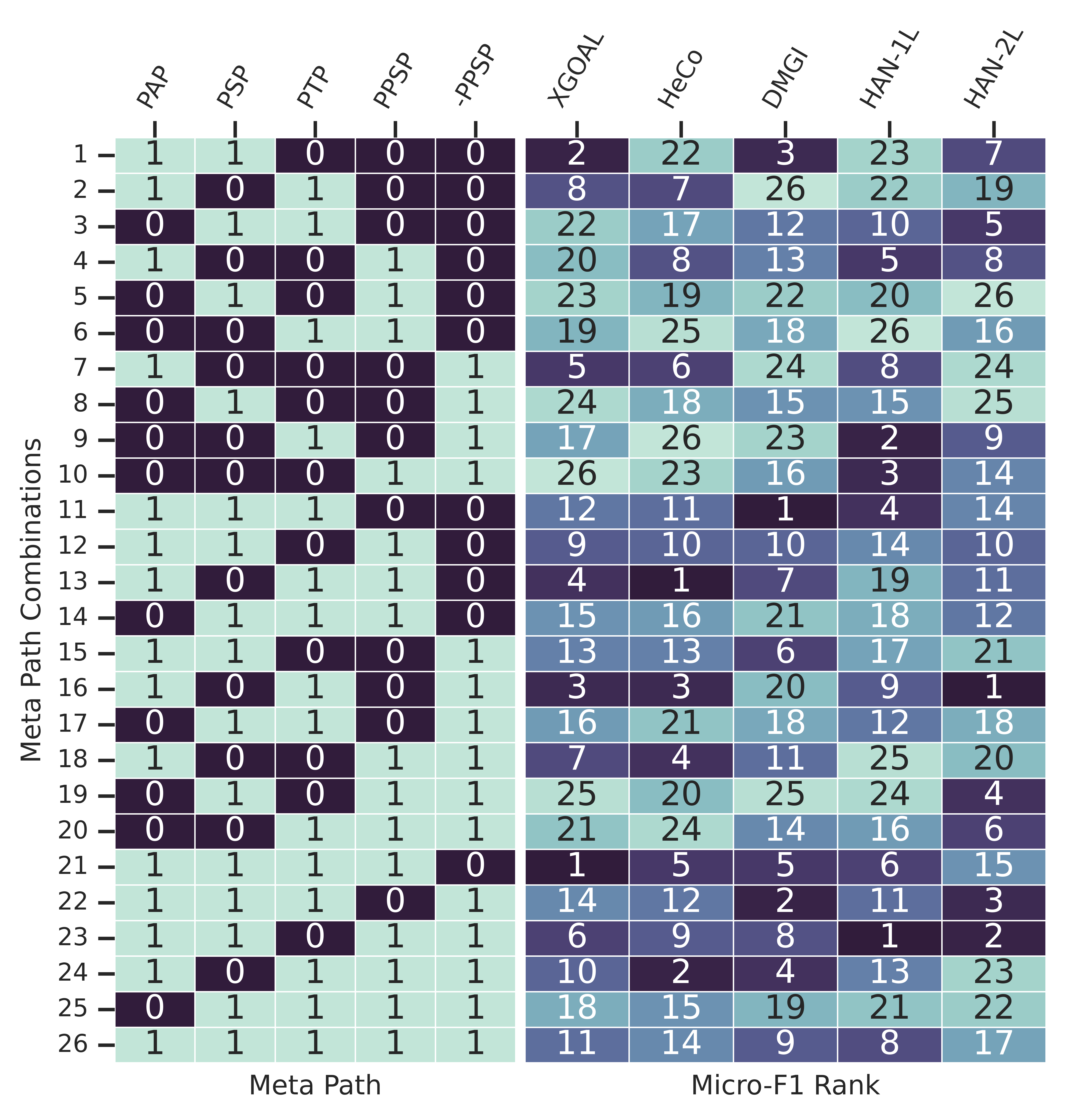} 
\caption{\small{We generated a total of 26 distinct meta-path combinations using five predefined meta-paths: PAP, PSP, PTP, PPSP, and -PPSP. A flag "1" indicates the inclusion of a particular meta-path in the combination, whereas the absence of a meta-path is denoted by a flag "0". Each column on the right side of the table ranks the performance of these meta-path combinations for different models.}}
\label{fig:metapath_comb_ranking} 
\end{figure}

To explore the influence of meta-path combinations on HGNN performance, we conducted a detailed empirical study using the ACM dataset. We generated 26 distinct combinations from 5 predefined meta-paths and assessed the performance variations in HGNNs, evidenced by the standard deviation and min-max gap. The key findings, depicted in Figures~\ref{fig:metapath_std} and~\ref{fig:metapath_comb_ranking}, are summarized below:

\noindent (1) \textbf{Sensitivity to Meta-path Combinations.} 
Meta-path combinations critically affect HGNN performance. Variations in these combinations impact the structural configuration of meta-path sub-graphs, significantly influencing model outcomes, as evidenced by the substantial standard deviation and min-max gap shown in Figure~\ref{fig:metapath_std}. In extreme cases, improper combinations can lead to model failure. This challenge is more acute in SSL models due to the lack of downstream task feedback. Even proven meta-paths can cause dramatic performance deterioration if combined inappropriately. The sensitivity of HGNNs to these combinations is partly due to their responsiveness to topological changes and is further compounded by the low homophily ratios in meta-path sub-graphs~\cite{guo2023homophily} (referenced in Table~\ref{table:HIN homophility}), which exacerbates the issue in denser sub-graph structures.

\noindent (2) \textbf{Absence of Universal Optimal Combinations.}: Our study reveals that no single meta-path combination is optimal for all models. This absence of a universal `best' combination becomes a formidable challenge in SSL, where the lack of direct feedback from downstream tasks makes finding the ideal combination through exhaustive search impractical. The disparity between the effective combinations in supervised and unsupervised models further complicates this issue. This gap suggests that strategies successful in supervised learning may not directly translate to superior performance in unsupervised settings.

\noindent (3) \textbf{Naively adding more meta-path do not guarantee the best.}: Contrary to expectations, simply adding more meta-paths does not linearly improve HGNN performance. While certain meta-paths are essential, their impact varies across different models. In some instances, such as the comparison between `comb26' and the optimal `comb21' for X-GOAL, adding an extra meta-path resulted in decreased performance. Our analysis, illustrated in Figure 4, shows significant edge overlaps among meta-path sub-graphs. For example, `-PPSP' overlaps with over 50\% of every other meta-path sub-graph. Such overlaps cause an accumulation of redundant information, overshadowing valuable insights from less common structures.  What's worse, current semantic-level aggregation methods struggle to filter out this redundancy, indicating that increasing meta-path count is not a straightforward solution for performance enhancement. In some cases, it can even be counterproductive.

\begin{table}[tbp]
\centering
\begin{tabular}{@{}cllll@{}}
\hline
\multicolumn{1}{c}{Dataset} & Meta-path & HR(\%) & ACC(\%)      &  Edges      \\ \midrule
                            & PAP       & 81.45  & 87.33 ± 0.56 &  29767     \\
                           & PSP       & 64.03  & 66.72 ± 0.49 &  2217089    \\
 ACM                        & PTP       & 33.38  & 68.21 ± 0.14 &  9150595   \\
                            & PcPSP     & 60.62  & 68.21 ± 1.08 &  1933761   \\
                            & PrPSP     & 61.41  & 68.16 ± 1.28 &  1440299   \\ \midrule          
\end{tabular}
\caption{\small{Homophility rate (HR) in different meta-path sub-graph of ACM dataset. ACC represents the node classification accuracy of 2-layer GCN with ReLU activation.}}
\label{table:HIN homophility}
\end{table}

This study's insights emphasize the critical need for a methodological strategy capable of forging a robust meta-path perspective, while simultaneously mitigating the redundancies that emerge from the amalgamation of various meta-paths.

\section{The Proposed Model: \textsf{LAMP}}

In this section, we introduce \textsf{LAMP}, a Learnable Meta-Path Guided Adversarial Contrastive Learning method, detailed in Figure \ref{fig:LAMP}. \textsf{LAMP} leverages a dual-view approach: a high-order information-rich meta-path view, processed by \textsf{LMA}, and a locally-focused network schema view. The essence of \textsf{LAMP} lies in its integration of diverse meta-path sub-graphs into a single, comprehensive meta-path sub-graph. To manage the inherent density of this integrated view, \textsf{LMA} – a meta-path guided learnable edge-pruning strategy – is employed. \textsf{LAMP}'s aim is to effectively retain essential sparsity for contrastive learning and reduce redundant information across the network schema and integrated meta-path views, enhancing node consistency across these views via an advanced adversarial training regime.

\begin{figure}[tbp] 
\centering 
\includegraphics[width=0.8\linewidth]{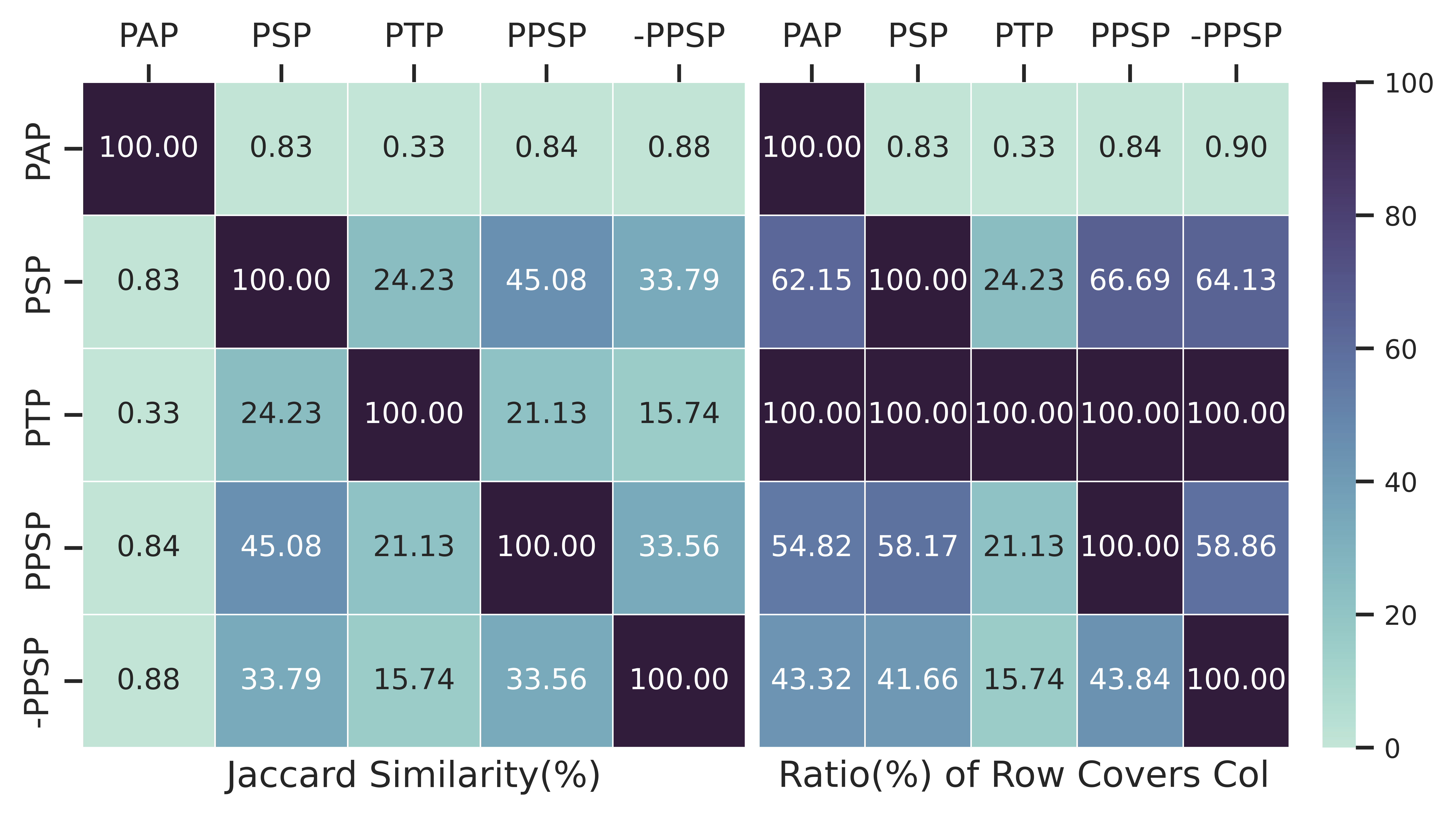} 
\caption{\small{We calculated Jaccard Similarity and coverage ratio based on meta-path instances (edges) in meta-path sub-graphs.}}
\label{fig:metapath_similarity}
\end{figure}

\begin{figure}[tbp]
\centering
\includegraphics[width=0.6\linewidth]{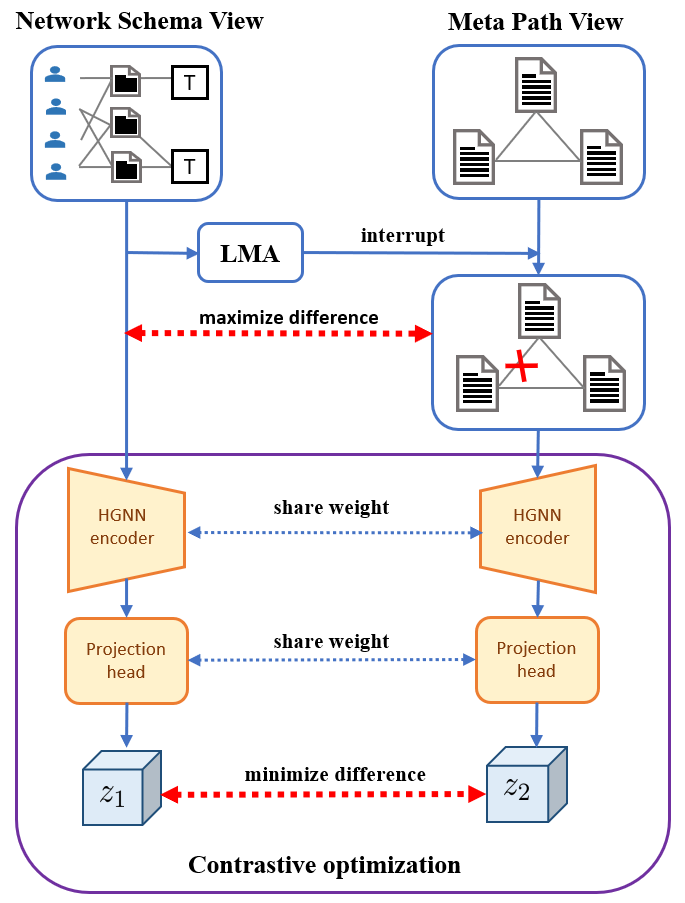}
\caption{\small{Overall architecture of the proposed \textsf{LAMP} model. \textsf{LAMP} processes network schema view $G$ and meta-path graph $t(\hat{G})$, which supply local and high-order information, respectively. The adversarial training mechanism is aplied to enhance the robustness of the meta-path view, alongside the contrastive optimization strategy employed to minimize the discrepancy between the two views.}}
\label{fig:LAMP}
\end{figure}

\subsection{Problem Formulation}
Given a HIN \( \mathcal{G} = (\mathcal{V}, \mathcal{E}, \mathcal{A}, \mathcal{R}) \) denoted as $G$ for short and a set of meta-path \( \{\mathcal{P}\} \) with \( |\{\mathcal{P}\}| = n \), we define \( \{\mathcal{G}_{\mathcal{P}_1} \dots \mathcal{G}_{\mathcal{P}_n}\}\) as meta-path sub-graphs and \({\mathcal{\hat{G}_P}}\) denoted as $\hat{G}$ as the integrated meta-path sub-graph. We represent the encoding function with parameter $\theta$ as $f(\cdot)$ and the augmentation function with parameter \( \phi \) as \( t(\cdot) \). For simplicity, we denote the network schema view by $f_\theta(G)$ and the meta-path view as $f_\theta(\hat{G})$. The primary objective for contrastive learning is:
\begin{equation}
    \arg \max_{\theta} I(f_{\theta}(G), f_{\theta}(t_{\phi}(\hat{G}))),
\end{equation}
Then for the adversarial training which tries to increase the difficulty of getting agreement in contrastive learning, the objective is:
\begin{equation}
  \arg \min_{\phi} I(f_{\theta}(G), f_{\theta}(t_{\phi}(\hat{G}))),
\end{equation}
where \( I(X_1; X_2) \) represents the mutual information between random variables \( X_1 \) and \( X_2 \). The graph \( t_{\phi}(\hat{G}) = (\hat{V},\hat{E}) \) retains the nodes from \( \hat{G} \), but its edge set is a subset of \( \hat{E} \). The insight is, we are trying to make the contrast as strong as possible while the two different view still could reach an agreement, which has been proved to be a effective optimization in contrastive learning. To bridge the min-max procedure and address potential biases, we incorporate a learnable meta-path importance parameter \( \gamma \in \mathbb{R}^{1 \times |\mathcal{P}|} \), which shared by \( f(\cdot) \) and \( t(\cdot) \). Then we put all the objective together and the refined version is:
\begin{equation}
     \arg \max_{\theta} \min_{\phi} I(f_{\theta}(G), f_{\theta , \gamma}(t_{\phi , \gamma}(\hat{G})).
\end{equation}
Consistent with prior research \cite{suresh2021adversarial,guan2023dmmg}, we employ InfoNCE~\cite{InfoNCE} to approximate \( I(X_1; X_2) \), detailed further in Section~\ref{section:contrastive loss}. Regarding $\gamma$, the insight is that \( \gamma \) prioritizes longer meta-paths because the most straightforward strategy for $t_\phi(\cdot)$ to diminish the similarity between the two views is by preserving long meta-path instances in $\hat{G}$. Conversely, during the maximization phase, shorter meta-paths become more influential. This balanced strategy empowers \textsf{LAMP} to harness rich high-order information while discerning the value of different meta-paths.

\subsection{Integrated Sub-graph based meta-path view}
Given a batch of meta-path sub-graphs $\{\mathcal{G}_{\mathcal{P}_i}\} = \{(\mathcal{V}_{\mathcal{P}_i},\mathcal{E}_{\mathcal{P}_i})\}$ with $i=1, \cdots ,|\mathcal{P}|$, we amalgamate all of them into a singular sub-graph denoted as $\hat{G} = (\hat{V},\hat{E})$, to create the meta-path view. Here $\hat{V} = \cup_i \mathcal{V}_{\mathcal{P}_i}$ and $\hat{E} = \cup_i \mathcal{E}_{\mathcal{P}_i}$. As an illustration, Figure~\ref{fig:HG}(c) depicts an integrated sub-graph derived from three meta-path sub-graphs: PAP, PSP, and PAPAP. The edge $e_{12} = (0,1,1)$ emerges since it's absent in PAP but present in both PSP and PAPAP. For every edge $ (u,v) \in \hat{E}$, we assign a vector $e_{uv} =  (x_1, x_2, \cdots, x_{|\mathcal{P}|})$ to present its semantic information, where $x_i$ is set to $1$ if $(u,v) \in \{\mathcal{E}_{\mathcal{P}_i}\}$, otherwise $x_i$ is set to $0$. To further capture the semantic level information, we assign a learnable vector $\gamma \in \mathbb{R}^{1 \times |\mathcal{P}|}$ that quantifies the importance of each meta-path. In the message-passing phase, we utilize $\hat{e}_{uv} = \gamma \times e_{uv}$ as the edge embedding within the meta-path view. This approach ensures that overlaps between meta-path sub-graphs are mitigated, thereby curtailing redundant message passing and rendering the meta-path view more robust compared to prior methodologies. Nevertheless, this method can lead to a dense meta-path view, an aspect we address through the proposed \textsf{LAMP}, detailed in the subsequent section.

\label{LMA}
\begin{figure*}[tbp]
\centering
\includegraphics[width=0.98\textwidth]{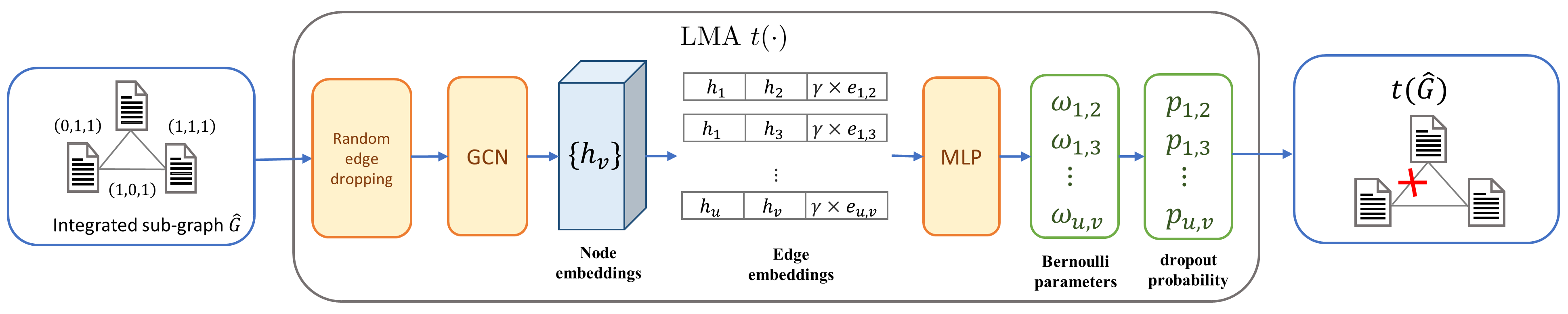}
\caption{\small{Overall architecture of LMA. To generate $t(\hat{G})$, LMA firstly accept the integrated sub-graph $\hat{G}$ then processes the node embedding of $\hat{G}$ using a two-layer GCN and combines these node embedding with the edge type embedding $ \hat{e}_{u,v}$ to form edge embedding and then fed into an MLP to determine Bernoulli parameters, which are ultimately converted to dropout probabilities utilizing the Gumbel-Max reparametrization trick.}}
\label{fig:LMA}
\end{figure*}

\subsection{Learnable Meta-Path Guided Augmentation}
The dense links of the integrated sub-graph, while capturing all given meta-path sub-graphs, can pose challenges for graph contrastive learning, as sparser graphs tend to yield more favorable results~\cite{zhu2021empirical}. To address this issue, we introduce the Learnable Meta-Path Augmentation (LMA), a adversarial training based method aimed at learning a optimized edge prunning strategy. This ensures a sparser meta-view while overcoming manually-induced biases. LMA firstly applies a random edge dropping then a learned GCN based edge prunning strategy based according to node feature and semantic information. At the first stage, random droping will effectively decrease the graph complexity since with the growth of engaged meta-path, the integrated sub-graph will be denser and eventually approximate a compeleted which is not desirable. Besides, random dropping will provdie the LMA a dynamic integrated sub-graph in each epoch so that LMA could learn a more powerful edge-cutting strategy rather than fall into a sub-optimal solution specialized for a fixed input. 
For learning edge cutting strategy, each edge \( e \in \hat{E} \) is correlated with a Bernoulli random variable, described as \( p_e \sim \textrm{Bernoulli}(\omega_e) \). An edge will be present in \( t_{\phi}(\hat{G}) = (V,E) \) if \( p_e = 1 \) and excluded otherwise. 
In order to cutting off edges based on not only node features, but also semantic information, the weights \( \omega_e \) of the Bernoulli distribution are parameterized using an MLP that takes as input the concatenation of the node embeddings obtained from a \( K \)-layer HGNN augmenter on \( \hat{G} \) and the edge type embedding $\hat{e}_{uv}$. Thus, the edge representations can be expressed as:
\begin{equation}
\omega_e = \text{MLP}\left( [h_{u}^{K}; h_{z}^{K}; \hat{e}_{uv}] \right).
\end{equation}
For a seamless end-to-end training of \( t_{\phi}(\hat{G}) \), the binary nature of \( p_e \) is transformed into a continuous variable between [0,1] using the Gumbel-Max reparametrization trick~\cite{maddison2016concrete,jang2016categorical}. Specifically:
\begin{equation}
        p_e = \mathrm{Sigmoid}\left(\frac{\log(\delta) - \log(1-\delta) + \omega_e}{\tau}\right),
\end{equation}
where \( \theta \sim \text{Uniform}(0,1) \). As \( \tau \) converges to zero, \( p_e \) gravitates towards binary values, ensuring the gradient remains smooth and well-defined. Notably, this kind of edge pruning, underpinned by a stochastic graph model, has also been utilized to provide parameterized explanations of GNNs~\cite{luo2020parameterized}.\par
To curb LMA's tendency for aggressive edge pruning, a regularization term \( \lambda_{reg} \frac{\sum_{e \in \hat{E}}\omega_e}{|\hat{E}|} \) is incorporated into the objective function. The hyper-parameter \( \lambda_{reg} \) dictates the quantity of retained edges. Without this regulation, LMA might opt for an extreme strategy of eliminating all edges to minimize the mutual information between \( G \) and \( t_\phi(\hat{G})\), which is counterproductive. This regularization ensures an edge ratio is maintained in \( t_\phi(\hat{G})\) to keep sufficient information for contrastive learning. The refined objective is:
\begin{equation}
    \arg \max_{\theta} \min_{\phi} \left( I(f_{\theta}(G), f_{\theta , \gamma}(t_{\phi , \gamma}(\hat{G})) - \lambda_{reg} \frac{\sum_{e \in \hat{E}} \omega_e}{|\hat{E}|} \right).
    \label{eq:objective with regulation}
\end{equation}

Of note, the meta-path importance $\gamma$ offers a holistic perspective for both $f_{\phi,\gamma}(\cdot)$ and $t_{\phi,\gamma}(\cdot)$. While $t_{\phi,\gamma}(\cdot)$ strives to maximize divergence from the network schema view, it places a premium on longer meta-paths. This is because, in contrast to the meta-path view, the network schema primarily harbors single-hop information. Conversely, during the agreement maximization phase, shorter meta-paths become more salient contributors by $f_{\phi,\gamma}(\cdot)$.

\subsection{Network Schema view}

In the network schema view, for a given node \( i \), we initiate the process by employing a type-specific multilayer perceptron (MLP), denoted as \( MLP^{\mathcal{A}(i)} \), to transform the features \( x_i \) of node \( i \) into a unified feature space. This transformation is represented as follows:
\begin{equation}
    h_i^{(0)} = MLP^{\mathcal{A}(i)}(x_i).
    \label{eq:objective with regulation_v1}
\end{equation}
Here, \( \mathcal{A}(i) \) presents the type of node \( i \). Subsequently, we incorporate one-hot encoding to represent the semantic information of various relations. This encoded information, along with the node features, is input into a unified HGNN encoder. The specifics of this unified HGNN encoder, which operates irrespective of node types while preserving edge type embeddings, will be elaborated in the following section.It is important to note that both the network schema view and the meta-path view utilize the same HGNN encoder, denoted as \( f_{\theta,\gamma} \). However, a key distinction lies in the treatment of the parameter \( \gamma \): it remains frozen in the network schema view, whereas gradients are enabled for \( \gamma \) in the meta-path view, allowing for adaptability in encoding different types of information.

\subsection{Unified HGNN Encoder}
\label{HGNN}
In the context of \textsf{LAMP}, as outlined in eq~\ref{eq:objective with regulation}, it is crucial to employ a unified HGNN that can efficiently handle both the network schema view (heterogeneous graph) and the meta-path view (homogeneous graph). While an approach could involve two distinct HGNN encoders tailored for each view, such an architecture may be inappropriate for \textsf{LAMP}. The core concern is that distinct encoders might produce node embedding governed by entirely different parameter sets, making it extremely hard for \textsf{LAMP} to meaningfully minimize similarity based on topological information. Essentially, a unified HGNN encoder fosters a harmonious link between the two views, ensuring that embedding reflects inherent structural divergence rather than encoder bias.
\begin{equation}
\small
    \hat{\alpha} = \frac{\text{exp}(\text{LeakyReLU}(a^{T}[Wh_i \Arrowvert Wh_j \Arrowvert W_r r_\psi(\langle i,j \rangle)]))}{\sum_{k \in N_i}\text{exp}(\text{LeakyReLU}(a^{T}[Wh_i \Arrowvert Wh_k \Arrowvert W_r r_\psi( \langle i,k \rangle)]))}.
\end{equation}
\subsubsection{Node Residual:}Introducing pre-activation residual connections for nodes:
\begin{equation}
h_i^{(l)} = \sigma(\sum_{j \in N_i} \alpha_{ij}^{(l)}W^{(l)}h_{j}^{l-1}+W_{res}^{(l)}h_i^{(l-1)}).
\end{equation}
\subsubsection{Edge Residual:}Following the insights from Realformer~\cite{EdgeRes}, we add residuals to the attention scores:
\begin{equation}
    \alpha_{ij}^{(l)} = (1-\beta)\alpha_{ij}^{(l)} + \beta\alpha_{ij}^{(l-1)},
\end{equation}
with \( \beta \in [0,1] \) serving as a scaling factor.
In our framework, the representation of relationships between end nodes varies based on the view. For the network schema view, the function $r_\psi( \langle u,v \rangle)$ yields a one-hot vector encapsulating the relation between the nodes. Conversely, in the meta-path view, the relationship is captured by $r_\psi( \langle u,v \rangle) = \hat{e}_{uv}$ leveraging the embedded semantic information.  The transformation matrix $W_r$ is designed to align the dimension of edge embedding with that of node embedding. Uniquely within the HGNN encoder, $W_r$ is the sole parameter not shared across both the network schema and meta-path views. 

\subsection{Contrastive Optimization}
\label{section:contrastive loss}
The core of our approach involves utilizing the network schema view \(G\) and meta-path view \(\hat{G}\) for the contrastive learning mechanism. Both graphs are fed into an HGNN followed by an MLP with a single hidden layer, mapping them into a space where the contrastive loss is computed:
\begin{align}
     z_{i}^{G,\textrm{proj}} & = W^{(2)}\sigma (W^{(1)}z_{i}^{G}+b^{(1)})+b^{(2)}, \\
     z_{i}^{\hat{G},\textrm{proj}} & = W^{(2)}\sigma (W^{(1)}z_{i}^{\hat{G}}+b^{(1)})+b^{(2)},
\end{align}
where \(\sigma\) denotes the Leaky Relu function. The parameters are shared between the two views' embedding. \par
Adopting the strategy introduced in HeCo, we generate high-quality positive and negative pairs. We introduce a connectivity vector \(C_i(j)\), which represents the connectivity between nodes based on the number of meta-path instances connecting them. 
\begin{equation}
    C_i(j) = \sum_{n = 1}^{|\mathcal{P}|} \mathbbm{1} (j \in N_i^{\mathcal{P}_n}),
\end{equation}
where $\mathbbm{1}(\cdot)$ represents the indicator function. Following this, we establish positive and negative samples by applying a threshold to the sorted node connectivity using $T_{pos}$. The intuition here is that node pairs with higher connectivity are more likely to belong to the same class. The contrastive loss for node $i$ can be defined as follows:
\begin{equation}
    \mathcal{L}_i = -\log \frac{\sum_{j \in Pos_i}\exp(sim(z_{i}^{G,\textrm{proj}},z_{j}^{\hat{G},\textrm{proj}})/\tau)}{\sum_{k \in Pos_i \cup Neg_i}\exp(sim(z_i^{G,\textrm{proj}},z_{k}^{\hat{G},\textrm{proj}})/\tau)},
\end{equation}
where \(sim(u,v)\) represents the cosine similarity between vectors \(u\) and \(v\), and \(\tau\) is the temperature parameter. The final objective aggregates the contrastive losses for all nodes:
\begin{equation}
      \mathcal{J} = \frac{1}{|V|}\sum_{i \in V}\mathcal{L}_i. 
\end{equation}
For downstream tasks, embedding from \(z^{\hat{G}}\) from the meta-path view is employed. Throughout the training process, a two-step approach is implemented for each epoch. For every epoch, in the first step, parameters within \textsf{LMP} are frozen, and we train the HGNN by minimizing the contrastive loss. Subsequently, in the second step, HGNN parameters are frozen while \textsf{LAMP} is trained with the objective of maximizing the contrastive loss.

\begin{table}[htbp]\centering
\begin{tabular}{ccccccc}
\hline
Dataset  & Nodes  & NodeTypes & Edges   & EdgeTypes & Target & Classes \\
\hline
DBLP     & \num{26128}  & 4         & \num{239566}  & 6         & author & 4       \\
IMDB     & \num{21420}  & 4         & \num{86642}   & 6         & movie  & 5       \\
ACM      & \num{10942}  & 4         & \num{547872}  & 8         & paper  & 3       \\
Freebase & \num{180098} & 8         & \num{1057688} & 36        & book   & 7      \\
\hline
\end{tabular}
\caption{\small{The statistics of the datasets}}
\label{table:dataset}
\end{table}

\section{Experimental Evaluation}
\begin{table*}[btp]
\resizebox{\linewidth}{!}{
\begin{tabular}{@{}cccccccccc@{}}
\toprule
\multicolumn{2}{c}{Dataset}        & \multicolumn{2}{c}{DBLP}                  & \multicolumn{2}{c}{IMDB}                  & \multicolumn{2}{c}{ACM}                   & \multicolumn{2}{c}{FreeBase}              \\ \midrule
Methods            & Training Data & Micro-F1            & Macro-F1            & Micro-F1            & Macro-F1            & Micro-F1            & Macro-F1            & Micro-F1            & Macro-F1            \\ \midrule
GCN                & X,A,P,Y       & 90.84±0.32          & 91.47±0.34          & 57.88±1.18          & 64.82±0.64          & 92.17±0.24          & 92.12±0.23          & 27.84±3.13          & 60.23±0.92          \\
RGCN               & X,A,Y         & 91.52±0.50          & 92.07±0.50          & 58.85±0.26          & 62.05±0.15          & 91.55±0.74          & 91.41±0.77          & 46.78±0.77          & 58.33±1.57          \\
HAN                & X,A,P,Y       & 91.67±0.49          & 92.05±0.62          & 57.74±0.96          & 64.63±0.58          & 90.89±0.43          & 60.79±0.43          & 21.31±1.68          & 54.77±1.4           \\
GTN                & X,A,Y         & 93.52±0.55          & 93.97±0.54          & 60.47±0.98          & 65.14±0.45          & 91.31±0.70          & 91.20±0.71          & OOM                 & OOM                 \\
HGT                & X,A,Y         & 93.01±0.23          & 93.49±0.25          & 63.00±1.19          & 67.20±0.57          & 91.12±0.76          & 91.00±0.76          & 29.28±2.52          & 60.51±1.16          \\
GAT                & X,A,P,Y       & 93.83±0.27          & 93.39±0.30          & 58.94±1.35          & 64.86±0.43          & 92.26±0.94          & 92.19±0.93          & 40.73±2.58          & 65.26±0.45          \\
\hdashline[1pt/0.5pt]
Mp2vec             & A,P           & 90.25±0.10          & 91.17±0.10          & 41.45±1.60          & 42.46±1.70          & 61.13±0.40          & 62.72±0.30          & 55.94±0.7           & 58.74±0.80          \\
DGI                & X,A,P         & 89.19±0.90          & 90.35±0.80          & 46.13±0.30          & 47.21±0.90          & 80.03±3.30          & 80.15±3.20          & 53.81±1.10          & 57.96±0.70          \\
DMGI               & X,A,P         & 89.46±0.60          & 90.66±0.50          & 47.49±1.40          & 61.97±1.30          & 87.97±0.40          & 87.82±0.50          & 52.10±0.70          & 56.69±1.20          \\
X-GOAL             & X,A,P         & 83.00±0.25          & \underline{91.90±0.22}          & 57.43±0.50          & 58.14±0.62          & \underline{91.22±0.10}          & \underline{91.26±0.17}          & 58.44±1.10          & 57.91±1.10          \\
HeCo               & X,A,P         & \underline{90.64±0.30}          & 91.59±0.20          & \underline{58.07±0.50}          & \underline{59.13±0.60}         & 89.04±0.50          & 88.71±0.50          & \underline{60.13±1.30}         & \underline{62.24±1.60}        \\
\midrule
\textbf{LAMP}      & X,A           & \textbf{92.44±0.32} & \textbf{92.22±0.30} & \textbf{61.85±0.39} & \textbf{62.19±0.50} & \textbf{91.35±0.50} & \textbf{91.27±0.50} & \textbf{61.32±1.20} & \textbf{64.13±1.20} \\ 

\bottomrule
\end{tabular}
}
\caption{\small{Quantitative results on node classification, detailing accuracy percentages and standard deviations. The second column specifies the training data available for each method, where \(X\), \(A\), \(P\), and \(Y\) correspond to node features, the adjacency matrix, optimal meta-path combination, and labels, respectively. The best and second best performance for unsupervised models is highlighted in \textbf{boldface} and \underline{underline}. Instances where the computation surpassed the memory constraints of a 200GB CPU are marked as "OOM".}}
\label{table:node classification}
\end{table*}

\subsection{Experimental Setup}

\subsubsection{Datasets:} In our study, we leveraged the HGB benchmark~\cite{HGB}, which includes four diverse HIN datasets detailed in Table \ref{table:dataset}. The \textbf{DBLP} dataset~\cite{MAGNN} is sourced from the renowned DBLP bibliography website, focusing on a subset of computer science publications and featuring nodes such as authors, papers, terms, and venues. The \textbf{ACM} dataset~\cite{ACMdata} is also a citation network from the computer science domain. We utilized the \textbf{Freebase} knowledge graph~\cite{Freebasedata}, specifically a subgraph with around 1,000,000 edges across eight types of entities, in line with previous research methodologies~\cite{Freebasedata1}. Lastly, the \textbf{IMDB} dataset focuses on the IMDB movie database, particularly covering movie genres like Action, Comedy, Drama, Romance, and Thriller.

\subsubsection{Baselines and Implementation Details:} We compare \textsf{LAMP} with a diverse set of methods, including five {\textsf unsupervised} techniques: Mp2vec~\cite{metapath2vec}, DGI~\cite{DGI}, DMGI~\cite{DMGI}, X-GOAL~\cite{XGOAL} and HeCo~\cite{HeCo}, as well as six {\textit (semi-)supervised} ones: GAT~\cite{GAT,HGB}, GCN~\cite{GCN,HGB}, RGCN~\cite{RGCN}, HAN~\cite{HAN}, GTN~\cite{GTN}, HGT~\cite{HGT}. For Mp2vec, we configure parameters with 40 walks per node, a walk length of 100, and a window size of 5. For the meta-path selection, in the case of Mp2vec and DGI, we evaluate all meta-paths and report the best results; for all the other meta-path based methods we report the best performance with their optimal meta-path combination. Unless stated otherwise, default parameters are adopted from the original papers. For GCN and GAT, we employ the approach outlined in~\cite{HGB}, enriching the original HIN with additional meta-path instances based on selected meta-paths. Specifically, for GAT, we employ the same edge-type embedding technique in the attention mechanism as in HGB. 
For \textsf{LAMP}, without a selection of optimal combination, we engage all the pre-defined meta-paths to construct the integrated meta-path subgraph. We use Glorot initialization~\cite{glorot} with the Adam optimizer~\cite{adam}. The learning rate ranges from \(1 \times 10^{-4}\) to \(5 \times 10^{-2}\), and patience values for early stopping are set between 5 and 200. Dropout rates are adjusted between 0.1 and 0.5, with increments of 0.05. \textsf{LMA} utilizes a two-layer GCN and \textsf{LAMP} integrates a two-layer HGB for node embedding within its contrastive learning framework. For the randomly edeg dropping, we search the best parameter from 0.3 to 0.8. We fixed the embedding dimensions at 64 for all techniques. Experiments are conducted 10 times randomly, with average results reported. For datasets lacking attributes, nodes receive one-hot ID vectors.

\subsection{Node Classification}
 In node classification task, we leveraged learned node embeddings to train a linear classifier in a transductive setting, utilizing all available edges during training. The distribution of node labels was consistent across datasets: 24\% for training, 6\% for validation, and 70\% for testing. Classification performance was evaluated using Macro-F1 and Micro-F1 metrics, with results reported for the test set based on optimal validation performance (Table \ref{table:node classification}). Among all baseline methods, we report the best performance with their corresponding optimal meta-path combinations For \textsf{LAMP}, we report the performance with combination involving all the meta-path to demonstrate the robustness. Notably, \textsf{LAMP} consistently surpassed other unsupervised methods and showed remarkable efficacy against supervised models, particularly in sparser datasets like IMDB and Freebase. Crucially, \textsf{LAMP} operates without relying on an optimal meta-path combination, setting it apart from other methodologies. We also examined \textsf{LAMP}'s sensitivity to meta-path combinations (Figure \ref{fig:metapath_std}), demonstrating its superior stability and robustness, even in comparison to supervised approaches.

\subsection{Sensitivity of Meta-Paths}

To examine the sensitivity of various meta-path combinations, we conducted experiments on the ACM dataset. Our focus was to observe the variations and the min-max gap in Micro-F1 scores across all possible meta-path combinations. We considered the following candidate meta-paths: "PAP", "PSP", "PTP", "PPSP", and "-PPSP", which collectively form 26 distinct meta-path combinations, as illustrated in Figure~\ref{fig:metapath_comb_ranking}. It is important to note that methods like Mp2vec and DGI were excluded from these experiments, as they are incompatible with all meta-path combinations due to their inherent design limitations and their inability to achieve state-of-the-art (SOTA) performance. The results of our experiments are presented in Table~\ref{table:robustness}. In these tests, \textsf{LAMP} demonstrated a significant outperformance over existing unsupervised methods and even surpassed some of the supervised learning methods in terms of Micro-F1 scores. Intriguingly, current state-of-the-art methods, including HeCo and Xgoal, exhibited substantial sensitivity to the choice of meta-path combinations. This finding underscores the importance of robust meta-path handling, especially in self-supervised learning contexts, and highlights the effectiveness of \textsf{LAMP} in addressing this challenge.

\begin{table}[htbp]
\centering
\small
\begin{tabular}{@{}c|c|c@{}}
\toprule
Methods       & Standard Deviation(\%) & Min-Max gap(\%) \\ \midrule
DMGI          & 5.46                   & 25.26           \\
XGOAL         & 7.01                   & 24.89           \\
HeCo          & 11.70                  & 36.69           \\
HAN-1Layer    & 3.95                   & 11.16           \\
HAN-2Layer    & 4.49                   & 20.82           \\
\midrule
\textbf{LAMP} & \textbf{2.07}          & \textbf{6.08}   \\ \bottomrule
\end{tabular}
\caption{\small{Quantitative results on Sensitivity of Meta-Paths}}
\label{table:robustness}
\end{table}

\subsection{Node Clustering}
In our experimental setup, we employ the K-means clustering algorithm for the learned node embedding. For performance evaluation, we utilize standard clustering metrics: normalized mutual information (NMI) and adjusted rand index (ARI). Recognizing the potential variability introduced by K-means due to its sensitivity to initialization, we execute the clustering process across ten independent runs and present the averaged outcomes in Table~\ref{table:node clustering}. Notably, the IMDB dataset is excluded from this evaluation, given its multi-dimensional label structure in HGB dataset. Furthermore, direct comparisons with supervised methodologies are omitted; these models have inherent access to label information during training and are optimized based on validation metrics. Empirical results underscore that \textsf{LAMP} consistently exhibits superior performance across datasets, reaffirming its effectiveness in the clustering context.

\begin{table}[htbp]
\centering
\small
\begin{tabular}{@{}c|cc|cc|cc@{}}
\hline
Datesets  & \multicolumn{2}{c|}{DBLP} & \multicolumn{2}{c|}{ACM} & \multicolumn{2}{c}{Freebase} \\ \midrule
Metrics  & NMI         & ARI        & NMI        & ARI        & NMI           & ARI          \\
\midrule
Mp2vec   & 73.55       & 77.70      & 48.43      & 34.65      & 16.47         & 17.32        \\
DGI      & 59.23       & 61.85      & 51.73      & 41.16      & 18.34         & 11.29        \\
DMGI     & 70.06       & 75.46      & 51.66      & 46.64      & 16.98         & 16.91        \\
X-GOAL   & 61.53       & 78.91      & 56.77      & 43.67      & 18.67         & 17.44        \\
HeCo     & 74.51       & 80.17      & 56.87      & 56.94      & 20.38         & 20.98        \\ \midrule
\textbf{LAMP} & \textbf{77.13} & \textbf{82.73} & \textbf{58.45} & \textbf{59.12} & \textbf{23.44} & \textbf{24.38}        \\ \bottomrule
\end{tabular}
\caption{\small{Quantitative results on node clustering.}}
\label{table:node clustering}
\vspace{-2.5em}
\end{table}

\subsection{Ablation Study}


This section evaluates two distinct variants: $\text{\textsf{LAMP}}_{\textrm{w.o.mp}}$ (referred to as $\text{\textsf{LAMP}}_{var1}$) and $\text{\textsf{LAMP}}_{\textrm{w.o.unifiedHGNN}}$ (referred to as $\text{\textsf{LAMP}}_{var2}$). For the $\text{\textsf{LAMP}}_{var1}$ version, we freeze the parameter $\gamma$ to cancel out the effect of meta-path importance during LMA learning. The intent behind this is to examine the role of meta-path importance in bridging local and high-order information. On the other hand, $\text{\textsf{LAMP}}_{var2}$ replaces the unified HGB encoder with the meta-path and network-schema encoders from HeCo. Within this setup, the meta-path view is processed using the HAN~\cite{HAN} attention mechanism, while a standard GCN tackles the original HIN. For the meta-path view, the \textsf{LMA} edge-pruning technique is applied to each individual meta-path sub-graph.\par
Table \ref{table:node classification} illustrates that both $\text{\textsf{LAMP}}_{var1}$ and $\text{\textsf{LAMP}}_{var2}$ suffer a considerable decline in performance. (1) Lacking the meta-path importance $\gamma$, $\text{\textsf{LAMP}}_{var1}$ struggles to harness sufficient overall structural data. It primarily emphasizes local details based on node attributes. Similarly, without the guidance of meta-path importance $\gamma$, LMA tends to prioritize lengthy meta-paths, and neglect potentially valuable shorter meta-paths. The resultant effect weakens $\text{\textsf{LAMP}}_{var1}$'s capability to bridge local and high-order information. This underscores that the guidance from meta-path importance is crucial for the \textsf{LAMP} model. (2) For $\text{\textsf{LAMP}}_{var2}$ , employing separate HGNN encoders for the two views might have been effective in HeCo, but it does not work for \textsf{LAMP}. As shown in Table\ref{table:abalation study}, $\text{\textsf{LAMP}}_{var2}$ lags behind in performance across all datasets.Using disparate HGNN encoders inherently amplifies the differences in embedding produced by the two views, even when the target node attributes remain consistent across both views. This introduces a dilemma for \textsf{LMA}, making it challenging to determine which edges to prune, as the two views already appear distinct. This inconsistency can destabilize the model, increasing the risk of training collapse.

\begin{table}[tbp]
\resizebox{\linewidth}{!}{
\begin{tabular}{@{}ccccccccc@{}}
\toprule
Dataset        & \multicolumn{2}{c}{DBLP}                  & \multicolumn{2}{c}{IMDB}                  & \multicolumn{2}{c}{ACM}                   & \multicolumn{2}{c}{FreeBase}              \\ \midrule
Methods        & Micro-F1            & Macro-F1            & Micro-F1            & Macro-F1            & Micro-F1            & Macro-F1            & Micro-F1            & Macro-F1            \\ \midrule
$\text{LAMP}_{var1}$ & 71.05          & 71.12          & 34.98         & 34.05          & 73.85          & 73.06          & 32.06          & 31.12         \\
$\text{LAMP}_{var2}$ & 86.33          & 87.27         & 53.40          & 54.20          & 84.54          & 84.75          & 49.51          & 50.04          \\
\textbf{LAMP}       & \textbf{92.44} & \textbf{92.22} & \textbf{61.85} & \textbf{62.19} & \textbf{91.35} & \textbf{91.27} & \textbf{61.32} & \textbf{64.13} \\ \bottomrule
\end{tabular}
}
\caption{\small{Quantitative results with two \textsf{LAMP} variants.}}
\label{table:abalation study}
\end{table}

\subsection{Analysis of Hyper-parameters}
In this section, we examine our model's sensitivity to two critical hyper-parameters: the threshold for positive samples $T_{pos}$ and the regulation term $\lambda_{reg}$, which determines the proportion of retained edges in LMA. Node classification on the ACM and DBLP datasets is evaluated, with both Macro-F1 and Micro-F1 scores presented.

\subsubsection{Analysis of $T_{pos}$}
The threshold $T_{pos}$ controls the number of positive samples. We vary its value to observe its impact on performance, as shown in Figure~\ref{fig:hyper-paremeter}(a) and Figure~\ref{fig:hyper-paremeter}(b). As $T_{pos}$ increases, performance initially improves before declining. The optimal thresholds are determined to be 7 for DBLP and 8 for ACM. These performance trends are consistent across both datasets.
\subsubsection{Analysis of $\lambda_{reg}$}
\begin{figure}[tbp]
\centering
\includegraphics[width=0.8\linewidth]{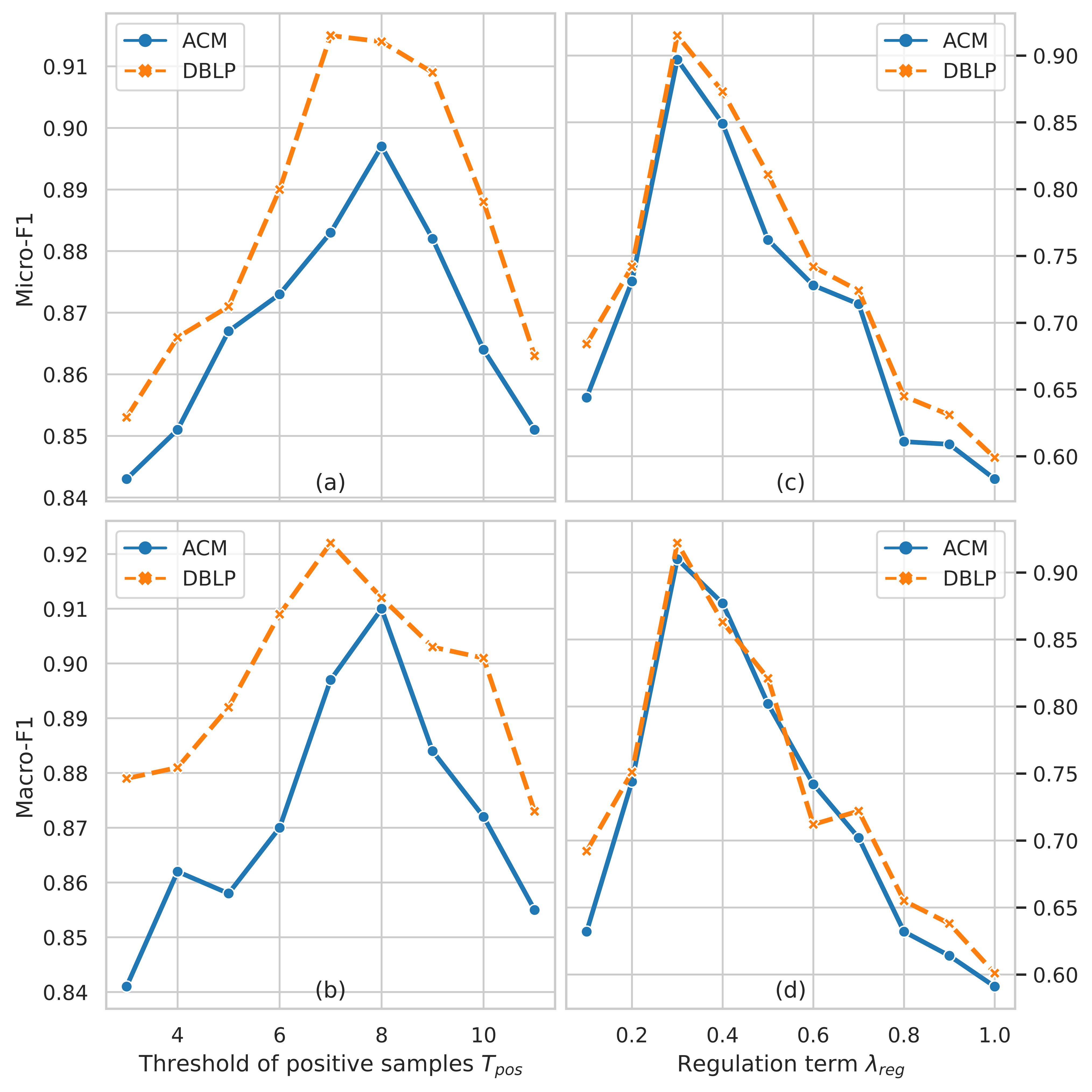}
\caption{\small{Impact of $T_{pos}$ and $\lambda_{reg}$ on performance.}}
\label{fig:hyper-paremeter}
\end{figure}

Our exploration also considers the consequences of adjusting $\lambda_{reg}$, which governs the fraction of edges retained by LMA. Results are presented in Figure~\ref{fig:hyper-paremeter}(c) and Figure~\ref{fig:hyper-paremeter}(d). For both DBLP and ACM datasets, $\lambda_{reg}$=0.3 yields peak performance, preserving approximately half of the meta-path view edges. Notably, raising $\lambda_{reg}$ beyond 0.5 results in the preservation of 70\%-80\% of edges. This excessive retention introduces redundant data into the model, leading to diminished efficacy.

\section{Related Work}

\subsection{Heterogeneous Graph Contrastive Learning}
HGCL has rapidly evolved, effectively adapting contrastive learning techniques for heterogeneous graphs~\cite{DMGI,hdmi,HeCo,XGOAL,chen2023heterogeneous,zheng2022contrastive,zhu2022structure}. Standard HGCL approaches involve creating multiple graph views via meta-path or network-schema based augmentations, followed by representation learning through contrasting positive and negative samples. DMGI~\cite{DMGI}, for instance, contrasts the original network with its corrupted counterpart for each meta-path view, integrating a consensus regularization for meta-path fusion. HeCo~\cite{HeCo} introduces two augmentation techniques—meta-path sub-graph view and network schema view—and minimizes the inter-view information entropy using personalized pairwise InfoNCE. HDMI~\cite{hdmi} and XGOAL~\cite{XGOAL} are advanced versions of DGMI. HDMI improved semantic attention via high-order mutual information, XGOAL proposed a stronger positive and negative samples generating strategy, and node embeddings are obtained by simply average pooling over these layer-specific embeddings. 
CPT-HG~\cite{CPT-HG} presents a pre-training model grounded in contrastive learning by making sub-graphs derived from positive samples integrate randomly swapped nodes from the negative set.

\subsection{HGNNs applications in IR}

In recent years, heterogeneous graph neural networks (HGNNs) as general extension of homogeneous graph \cite{huang2021towards,wang2022single,huang2023graph,wang2022pruning,zhang2024graph,wang2024heterophilic,han2024node} have risen to prominence as a pivotal tool in information retrieval (IR), adept at extracting rich structural and semantic information from heterogeneous graphs. This capability has led to their widespread application across various IR domains, including search engines, recommendation systems, and question-answering systems, among others. In the context of search engines and matching, Chen et al. \cite{DBLP:conf/mm/ChenWC00P22} innovated a cross-modal retrieval method utilizing heterogeneous graph embeddings. This method adeptly preserves cross-modal information, overcoming the limitations of traditional approaches that often lose modality-specific details. Similarly, Guan et al. \cite{guan2022personalized} addressed fashion compatibility modeling by integrating user preferences and attribute entities within a meta-path-guided HGNN framework. Additionally, Yuan et al. \cite{10.1145/3397271.3401159} introduced the Spatio-Temporal Dual Graph Attention Network (STDGAT) for intelligent query-Point of Interest (POI) matching in location-based services. By leveraging semantic representation, dual graph attention, and spatiotemporal factors, STDGAT enhances matching accuracy, even with partial query keywords.The domain of recommendation systems has also seen significant advancements through the application of HGNNs. Cai et al. \cite{cai2023user} proposed an inductive heterogeneous graph neural network (IHGNN) model tailored for cold-start recommendation scenarios, addressing the challenge of sparse user attribute data. Pang et al. \cite{pang2022heterogeneous} developed a personalized session-based recommendation method using heterogeneous global graph neural networks (HG-GNN), which effectively captures user preferences from both current and historical sessions. Moreover, Song et al. \cite{10.1145/3524618} presented a self-supervised, calorie-aware heterogeneous graph network (SCHGN) for food recommendations, integrating user preferences and ingredient relationships to enhance the recommendation quality.In the arena of question-answering systems, HGNNs have garnered considerable attention. Feng et al. \cite{feng2022multi} proposed a document-entity heterogeneous graph network (DEHG) that integrates structured and unstructured information sources for multi-hop reasoning in open-domain question answering. Furthermore, Gao et al. \cite{gao2022heteroqa} introduced HeteroQA, employing a question-aware heterogeneous graph transformer to assimilate multiple information sources from user communities, enriching the question-answering process.

\section{Conclusion}
Our study reveals the sensitivity of existing methodologies to meta-path combinations in unsupervised heterogeneous graph neural networks. To address this challenge, we introduce \textsf{LAMP}, a meta-path-guided adversarial approach for Heterogeneous Graph Contrastive Learning (HGCL). \textsf{LAMP} excels in capturing local and high-order structural information through dual views and Learnable Meta-Path guided augmentation (LMA) with an HGNN. Empirical tests across various datasets showcase \textsf{LAMP}'s superiority over existing unsupervised models and competitive performance even with supervised models. \textsf{LAMP} holds great potential for future heterogeneous graph contrastive learning research.

\bibliography{reference}
\bibliographystyle{plain}



\end{document}